%% file: emnlp2020.tex
\documentclass[11pt,a4paper]{article}
\usepackage[hyperref]{emnlp2020}
\usepackage{times}
\usepackage{amsmath}
\usepackage{amssymb}
\usepackage{amsthm}
\usepackage{mathtools}
\usepackage{latexsym}
\usepackage{makecell}

\usepackage{microtype}

\aclfinalcopy 
\def\aclpaperid{2514} 

    \newcommand{\argmax}{\mathop{\mathrm{argmax}}\limits}  
\DeclareMathOperator{\EX}{\mathbb{E}}

\title{SPARTA: Efficient Open-Domain Question Answering via Sparse Transformer Matching Retrieval}

\author{
  Tiancheng Zhao$^{1}$, Xiaopeng Lu$^{2}$ and Kyusong Lee$^{1}$ \\
  SOCO Inc. \\
  $^{1}$\texttt{\{tianchez,kyusongl\}@soco.ai} \\
  Language Technologies Institute, Carnegie Mellon University\\
  $^{2}$\texttt{xiaopen2@andrew.cmu.edu}
  }

\begin{document}
\maketitle
\begin{abstract}
{
We introduce SPARTA, a novel neural retrieval method that shows great promise in performance, generalization, and interpretability for open-domain question answering. Unlike many neural ranking methods that use dense vector nearest neighbor search, SPARTA learns a sparse representation that can be efficiently implemented as an Inverted Index. The resulting representation enables scalable neural retrieval that does not require expensive approximate vector search and leads to better performance than its dense counterpart. We validated our approaches on 4 open-domain question answering (OpenQA) 
tasks and 11 retrieval question answering (ReQA) tasks. SPARTA achieves new state-of-the-art results across a variety of open-domain question answering tasks in both English and Chinese datasets, including open SQuAD, Natural Question, CMRC and etc. Analysis also confirms that the proposed method creates human interpretable representation and allows flexible control over the trade-off between performance and efficiency.~\footnote{Work done during Lu's internship at SOCO.}
}

\end{abstract}

\section{Introduction}
\input{intro}

\section{Related Work}
\input{related_work}

\section{Proposed Method}
\input{method}

\section{OpenQA Experiments}
\input{oqa_exp}
\input{oqa_results}

\section{Retrieval QA Experiments}
\input{reqa_exp}
\input{reqa_results}

\section{Model Analysis}
\input{analysis}



\section{Conclusion}
In short, we propose SPARTA, a novel ranking method, that learns sparse representation for better open-domain QA. Experiments show that the proposed framework achieves the state-of-the-art performance for 4 different open-domain QA tasks in 2 languages and 11 retrieval QA tasks. This confirm our hypothesis that token-level interaction is superior to sequence-level interaction for better evidence ranking. Analyses also show the advantages of sparse representation, including interpretability, generalization and efficiency.

Our findings also suggest promising future research directions. The proposed method does not support multi-hop reasoning, an important attribute that enables QA systems to answer more complex questions that require collecting multiple evidence passages. Also, current method only uses a bag-of-word features for the query. We expect further performance gain by incorporating more word-order information.

\newpage
\bibliography{emnlp2020}
\bibliographystyle{acl_natbib}

\appendix

\section{Appendices}
\label{sec:appendix}
Size details of multi-domain ReQA task.
\begin{table}[ht]
\small
\centering
\begin{tabular}{p{0.1\textwidth}|p{0.12\textwidth}p{0.2\textwidth}} \hline
\textbf{Domain}                        & \textbf{\# of Query}  & \textbf{\# of Candidate Answer} \\ \hline
SQuAD         & 11,426  & 10,250                    \\
News          & 8,633        &38,199               \\
Trivia          & 8,149  & 17,845                        \\
NQ       & 1,772  &  7,020                \\
Hotpot          & 5,901      & 38,906                 \\ 
BioASQ   & 1,562  & 13,802                  \\
DROP                  & 1,513  & 2,488                       \\
DuoRC          & 1,568  & 5,241                  \\
RACE                  & 674  & 10,630                   \\
RE                   & 2,947  & 2,201                       \\
Textbook          &  1,503  & 14,831                  \\ \hline
\end{tabular}
\caption{Size of the evaluation test set for the 11 corpora included in MultiReQA.}
\label{tbl:reqa_size}
\end{table}


\end{document}

%% file: intro.tex
Open-domain Question Answering (OpenQA) is the task of answering a question based on a knowledge source. One promising approach to solve OpenQA is Machine Reading at Scale (MRS)~\cite{chen2017reading}. MRS leverages an information retrieval (IR) system to narrow down to a list of relevant passages and then uses a machine reading comprehension reader to extract the final answer span. This approach, however, is bounded by its pipeline nature since the first stage retriever is not trainable and may return no passage that contains the correct answer. 

To address this problem, prior work has focused on replacing the first stage retriever with a trainable ranker~\cite{chidambaram2018learning,lee2018ranking,wang2018r}. End-to-end systems have also been proposed to combine passage retrieval and machine reading by directly retrieving answer span~\cite{seo2019real,lee2019latent}. Despite of their differences, the above approaches are all built on top of the \textit{dual-encoder architecture}, where query and answer are encoded into fixed-size dense vectors, and their relevance score is computed via dot products. Approximate nearest neighbor (ANN) search is then used to enable real-time retrieval for large dataset~\cite{shrivastava2014asymmetric}. 

In this paper, we argue that the dual-encoder structure is far from ideal for open-domain QA retrieval. Recent research shows its limitations and suggests the importance of modeling complex queries to answer interactions for strong QA performance. \citet{seo2019real} shows that their best performing system underperforms the state-of-the-art due to query-agnostic answer encoding and its over-simplified matching function. \citet{humeau2019poly} shows the trade-off between performance and speed when moving from expressive cross-attention in BERT~\cite{devlin2018bert} to simple inner product interaction for dialog response retrieval. Therefore, our key research goal is to develop new a method that can simultaneously achieve expressive query to answer interaction and fast inference for ranking.

We introduce SPARTA (Sparse Transformer Matching), a novel neural ranking model. Unlike existing work that relies on a sequence-level inner product, SPARTA uses token-level interaction between every query and answer token pair, leading to superior retrieval performance. Concretely, SPARTA learns sparse answer representations that model the potential interaction between every query term with the answer. The learned sparse answer representation can be efficiently saved in an Inverted Index, e.g., Lucene~\cite{mccandless2010lucene}, so that one can query a SPARTA index with almost the same speed as a standard search engine and enjoy the more reliable ranking performance without depending on GPU or ANN search. 

Experiments are conducted on two settings: OpenQA~\cite{chen2017reading} that requires phrase-level answers and retrieval QA (ReQA) that requires sentence-level answers~\cite{ahmad2019reqa}. Our proposed SpartaQA system achieves new state-of-the-art results across 15 different domains and 2 languages with significant performance gain, including OpenSQuAD, Open Natural Questions, OpenCMRC and etc. 

Moreover, model analysis shows that SPARTA exhibits several desirable properties. First SPARTA shows strong domain generalization ability and achieves the best performance compared to both classic IR method and other learning methods in low-resources domains. Second, SPARTA is simple and efficient and achieves better performance than many more sophisticated methods. Lastly, it provides a human-readable representation that is easy to interpret.  In short, the contributions of this work include:
\begin{itemize}
    \item A novel ranking model SPARTA that offers token-level query-to-answer interaction and enables efficient large-scale ranking.
    \item New state-of-the-art experiment results on 11 ReQA tasks and 4 OpenQA tasks in 2 languages.  
    \item Detailed analyses that reveal insights about the proposed methods, including generalization and computation efficiency.
\end{itemize}

%% file: related_work.tex
The classical approach for OpenQA depends on knowledge bases (KB)s that are manually or automatically curated, e.g., Freebase KB~\cite{bollacker2008freebase}, NELL~\cite{fader2014open} etc. Semantic parsing is used to understand the query and computes the final answer~\cite{berant2013semantic,berant2014semantic}. However, KB-based systems are often limited due to incompleteness in the KB and inflexibility to changes in schema~\cite{ferrucci2010building}. 

A more recent approach is to use text data directly as a knowledge base. Dr.QA uses a search engine to filter to relevant documents and then applies machine readers to extract the final answer~\cite{chen2017reading}. It needs two stages because all existing machine readers, for example, BERT-based models ~\cite{devlin2018bert}, are prohibitively slow (BERT only processes a few thousands of words per second with GPU acceleration). Many attempts have been made to improve the first-stage retrieval performance~\cite{chidambaram2018learning,seo2019real,henderson2019convert,karpukhin2020dense,chang2020pre}. Yet, the information retrieval (IR) community has shown that simple word embedding matching do not perform well for ad-hoc document search compared to classic methods~\cite{guo2016deep,xiong2017end}. 

To increase the expressiveness of dual encoders, \citet{xiong2017end} develops kernel function to learn soft matching score at token-level instead of sequence-level. \citet{humeau2019poly} proposes Poly-Encoders to enable more complex interactions between the query and the answer by letting one encoder output multiple vectors instead of one vector. \citet{dhingra2020differentiable} incorporates entity vectors and multi-hop reasoning to teach systems to answer more complex questions. \cite{Lee2020ContextualizedSR} augments the dense answer representation with learned n-gram sparse feature from contextualized word embeddings, achieving significant improvement compared to the dense-only baseline. \citet{chang2020pre} explores various unsupervised pretraining objectives to improve dual-encoders' QA performance in the low-resources setting.


Unlike most of the existing work based-on dual-encoders, we explore a different path where we focus on learning sparse representation and emphasizes token-level interaction models instead of sequence-level. This paper is perhaps the most related to the sparse representations from~\cite{Lee2020ContextualizedSR}. However, the proposed approach is categorically different in the following ways. (1) it is stand-alone and does not need augmentation with dense vectors while keeping superior performance (2) our proposed model is architecturally simpler and is generative so that it will understand words that not appear in the answer document, whereas the one developed at~\cite{Lee2020ContextualizedSR} only models n-grams appear in the document.

%% file: method.tex
\begin{figure*}[t]
\centering
\includegraphics[width=16cm]{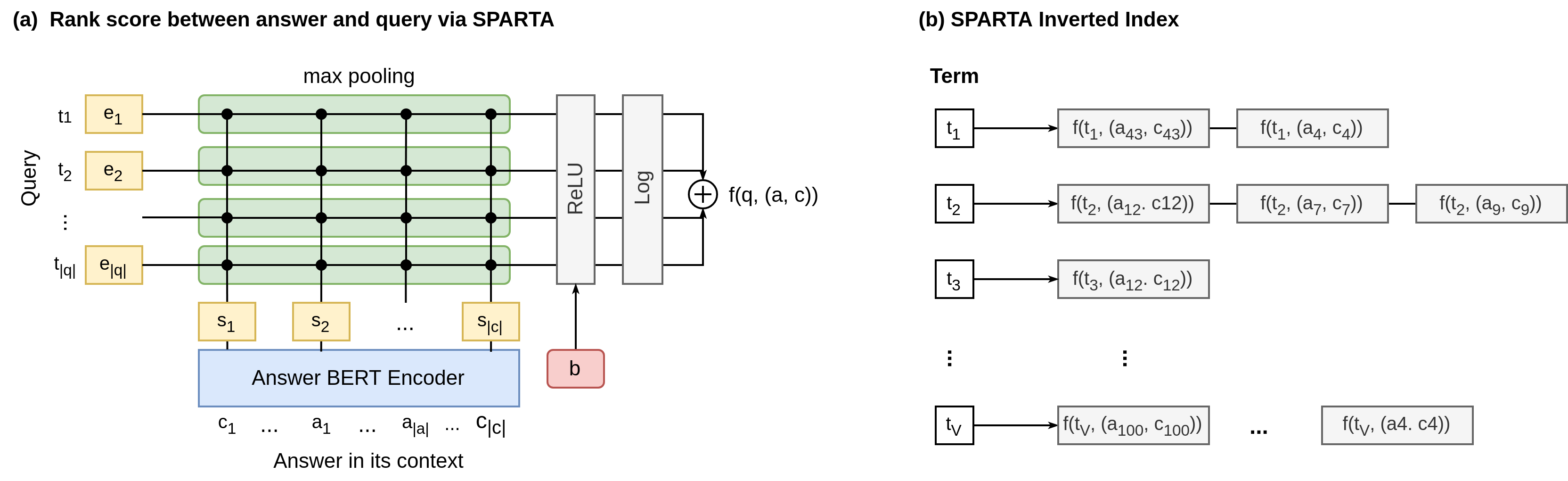}
\caption{SPARTA Neural Ranker computes token-level matching score via dot product. Each query terms' contribution is first obtained via max-pooling and then pass through ReLU and log. The final score is the summation of each query term contribution.}
\label{fig:overview}
\end{figure*}

\subsection{Problem Formulation}
First, we formally define the problem of answer ranking for question answering. Let $q$ be the input question, and $A=\{(a, c)\}$ be a set of candidate answers. Each candidate answer is a tuple $(a, c)$ where $a$ is the answer text and $c$ is context information about $a$. The objective is to find model parameter $\theta$ that rank the correct answer as high as possible, .i.e:
\begin{equation}
    \theta = \argmax_{\theta \in \Theta} \EX[p_{\theta}((a^*, c^*)|q)] 
\end{equation}

This formulation is general and can cover many tasks. For example, typical passage-level retrieval systems sets the $a$ to be the passage and leaves $c$ empty~\cite{chen2017reading,yang2019end}. The sentence-level retrieval task proposed at sets $a$ to be each sentence in a text knowledge base and $c$ to be the surrounding text~\cite{ahmad2019reqa}. Lastly, the phrase-level QA system sets $a$ to be all valid phrases from a corpus and $c$ to be the surrounding text~\cite{seo2019real}. This work focuses on the same sentence-level retrieval task~\cite{ahmad2019reqa} since it provides a good balance between precision and memory footprint. Yet note that our methods can be easily applied to the other two settings.

\subsection{SPARTA Neural Ranker}
In order to achieve both high accuracy and efficiency (scale to millions of candidate answers with real-time response), the proposed SPARTA index is built on top of two high-level intuitions.
\begin{itemize}
    \item Accuracy: retrieve answer with expressive embedding interaction between the query and answer, i.e., token-level contextual interaction. 
    \item Efficiency: create query agnostic answer representation so that they can be pre-computed at indexing time. Since it is an offline operation, we can use the most powerful model for indexing and simplify the computation needed at inference.
\end{itemize}

As shown in Figure~\ref{fig:overview}, a query is represented as a sequence of tokens $q=[t_1, ... t_{|q|}]$ and each answer is also a sequence of tokens $(a, c)= [c_1, .. a_1, .. a_{|a|}, c_{a+1}, ... c_{|c|}]$. We use a \textbf{non-contextualized} embedding to encode the query tokens to $e_i$, and a \textbf{contextualized} transformer model to encode the answer and obtain contextualized token-level embedding $s_j$:
\begin{align}
    \mathcal{E}(q) = [e_1, ... e_{|q|}] \quad \text{Query Embedding}&\\
    \mathcal{H}(a, c) = [s_1, ... s_{|c|}] \quad \text{Answer Embedding}&
\end{align}

Then the matching score $f$ between a query and an answer is computed by:
\begin{align}
    y_i = \text{max}_{j \in [1, |c|]} (e_i^T s_j) \quad \text{Term Matching}& \label{eq:term_match}\\ 
    \phi(y_i) = \text{ReLU}(y_i + b) \quad \text{Sparse Feature}& \label{eq:sparse}\\
    f(q, (a, c)) = \sum_{i=0}^{|q|}\log(\phi(y_i)+1) \quad \text{Final Score}&
\end{align}
where $b$ is a trainable bias. The final score between the query and answer is the summation of all individual scores between each query token and the answer. The logarithm operations normalize each individual score and weaken the overwhelmingly large term score. Additionally, there are two key design choices worth of elaboration.

\textbf{Token-level Interaction}
SPARTA scoring uses token-level interaction between the query and the answer. Motivated by bidirectional-attention flow~\cite{seo2016bidirectional}, relevance between every query and answer token pair is computed via dot product and max pooling in Eq.~\ref{eq:term_match}. Whereas in a typical dual-encoder approach, only sequence-level interaction is computed via dot product. Results in our experiment section show that fine-grained interaction is crucial to obtain significant accuracy improvement. Additionally, $s_j$ is obtained from powerful bidirectional transformer encoders, e.g. BERT and only needs to be computed at the indexing time. On the other hand, the query embedding is non-contextual, a trade-off needed to enable real-time inference, which is explained in Section~\ref{sec:infer} 

\textbf{Sparsity Control}
Another key feature to enable efficient inference and memory foot print is sparsity. This is achieved via the combination of $\log$, $\text{ReLU}$ and $b$ in Eq.~\ref{eq:sparse}. The bias term is used as a threshold for $y_i$. The $\text{ReLU}$ layer forces that only query terms with $y_i > 0$ have impact to the final score, achieving sparse activation. The $\log$ operation is proven to be useful via experiments for regularizing individual term scores and leads to better performance and more generalized representation.

\textbf{Implementation}
In terms of implementation, we use a pretrained 12-layer, 768 hidden size \textit{bert-base-uncased} as the answer encoder to encode the answer and their context~\cite{devlin2018bert}. To encode the difference between the answer sequence and its surrounding context, we utilized the segment embedding from BERT, i.e. the answer tokens have $\text{segment\_id}=1$ and the context tokens have$\text{segment\_id}=0$. Moreover, the query tokens are embedded via the word embedding from the bert-base-uncased with dimension 768.    

\subsection{Learning to Rank}
The training of SPARTA uses cross entropy learning-to-rank loss and maximizes Eq.~\ref{eq:loss}. The objective tries to distinguish between the true relevant answer $(a^+, c^+) $and irrelevant/random answers $K^-$ for each training query $q$:
\begin{equation}
    J = f(q, (a^+, c^+)) - \log\sum_{k \in K^-}e^{f(q, (a_k, c_k))}
\label{eq:loss}
\end{equation}
The choice of negative samples $K^-$ are crucial for effective learning. Our study uses two types of negative samples: 50\% of the negative samples are randomly chosen from the entire answer candidate set, and the rest 50\% are chosen from sentences that are nearby to the ground truth answer $a$. The second case requires the model to learn the fine-grained difference between each sentence candidate instead of only rely on the context information. The parameters to learn include both the query encoder $\mathcal{E}$ and the answer encoder $\mathcal{H}$. Parameters are optimized using back propagation (BP) through the neural network. 

\subsection{Indexing and Inference}
\label{sec:infer}
One major novelty of SPARTA is how one can use it for real-time inference. That is for a testing query $q=[t_0, ... t_{|q|}]$, the ranking score between $q$ and an answer is:
\begin{align}
    &\text{LOOKUP}(t, (a, c)) = \log(\text{Eq.~\ref{eq:sparse}}) \quad t \in V & \label{eq:index}\\
    &f(q, (a, c)) = \sum_{i=1}^{|q|}\text{LOOKUP}(t_i, (a, c)) \label{eq:infernece}&
\end{align}
Since the query term embedding is non-contextual, we can compute the rank feature $\phi(t, (a,c))$ for every possible term $t$ in the vocabulary $V$ with every answer candidate. The result score is cached in the indexing time as shown in Eq.~\ref{eq:index}. At inference time, the final ranking score can be computed via O(1) look up plus a simple summation as shown in Eq.~\ref{eq:infernece}.

More importantly, the above computation can be efficiently implemented via a Inverted Index~\cite{manning2008introduction}, which is the underlying data structure for modern search engines, e.g. Lucene~\cite{mccandless2010lucene} as shown in Figure~\ref{fig:overview}(b). This property makes it easy to apply SPARTA to real-world applications.

\subsection{Relation to Classic IR and Generative Models}
It is not hard to see the relationship between SPARTA and classic BM25 based methods. In the classic IR method, only the tokens that appeared in the answer are saved to the Inverted Index. Each term's score is a combination of \textit{Term Frequency} and \textit{Inverted Document Frequency} via heuristics~\cite{manning2008introduction}. On the other hand, SPARTA learns which term in the vocabulary should be inserted into the index, and predicts the ranking score directly rather than heuristic calculation. This enables the system to find relevant answers, even when none of the query words appeared in the answer text. For example, if the answer sentence is ``Bill Gates founded Microsoft", a SPARTA index will not only contain the tokens in the answer, but also include relevant terms, e.g. \textit{who}, \textit{founder}, \textit{entrepreneur} and etc.

SPARTA is also related to generative QA. The scoring between $(a, c)$ and every word in the vocabulary $V$ can be understood as the un-normalized probability of $\log p(q|a) = \sum_{i}^{|q|} \log p(t_i|a)$ with term independence assumption.
Past work such as \citet{lewis2018generative,nogueira2019doc2query} trains a question generator to score the answer via likelihood. However, both approaches focus on auto-regressive models and the quality of question generation and do not provide an end-to-end solution that enables stand-alone answer retrieval.

%% file: oqa_exp.tex
We consider an Open-domain Question Answering (OpenQA) task to evaluate the performance of SPARTA ranker. Following previous work on OpenQA \cite{chen2017reading, wang2019multi, xie2020distant}, we experiment with two English datasets: SQuAD~\cite{rajpurkar2016squad}, Natural Questions (NQ)~\cite{kwiatkowski2019natural}; and two Chinese datasets: CMRC~\cite{cui2018span}, DRCD~\cite{shao2018drcd}. For each dataset, we used the version of Wikipedia where the data was collected from. Preliminary results show that it is crucial to use the right version of Wikipedia to reproduce the results from baselines. We compare the results with previous best models.

System-wise we follow the 2-stage ranker-reader structure used in~\cite{chen2017reading}. 

\textbf{Ranker:}
We split all documents into sentences. Each sentence is treated as a candidate answer $a$. We keep the surrounding context words of each candidate answer as its context $c$. We encode at most 512 word piece tokens and truncate the context surrounding the answer sentence with equal window size. For model training, \textit{bert-base-uncased} is used as the answer encoder for English, and \textit{chinese-bert-wwm} is used for Chinese. We reuse the word embedding from corresponding BERT model as the term embedding. Adam~\cite{kingma2014adam} is used as the optimizer for fine-tuning with a learning rate 3e-5. The model is fine-tuned for at most 10K steps and the best model is picked based on validation performance.

\textbf{Reader:} We deploy a machine reading comprehension (MRC) reader to extract phrase-level answers from the top-K retrieved contexts. For English tasks, we fine-tune on \textit{span-bert}~\cite{joshi2020spanbert}. For Chinese tasks, we fine-tune on \textit{chinese-bert-wwm}~\cite{cui-etal-2020-revisiting}. Two additional proven techniques are used to improve performance. First, we use global normalization~\cite{clark2017simple} to normalize span scores among multiple passages and make them comparable among each other. Second, distant supervision is used. Concretely, we first use the ranker to find top-10 passages for all training data from Wikipedia corpus. Then every mention of the oracle answers in these contexts are treated as training examples. This can ensure the MRC reader to adapt to the ranker and make the training distribution closer to the test distribution ~\cite{xie2020distant}.

Lastly, evaluation metrics include the standard MRC metric: EM and F1-score.

\begin{itemize}
    \item Exact Match (EM): if the top-1 answer span matches with the ground truth exactly.
    \item F1 Score: we compute word overlapping between the returned span and the ground truth answer at token level.
\end{itemize}

%% file: oqa_results.tex
\subsection{OpenQA Results}
\begin{table}[ht!]
\centering
\begin{tabular}{p{0.32\textwidth}p{0.05\textwidth}p{0.05\textwidth}} 
\hline
\textbf{OpenSQuAD} &  &  \\ 
Model & F1 & EM \\ \hline
Dr.QA\cite{chen2017reading}  & - & 29.8  \\ 
R\textsuperscript{3} \cite{wang2018r} & 37.5 & 29.1 \\
Par. ranker \cite{lee2018ranking} & - & 30.2 \\
MINIMAL \cite{min2018efficient} & 42.5 & 32.7 \\
DenSPI-hybrid \cite{seo2019real} & 44.4 & 36.2 \\
BERTserini \cite{yang2019end} & 46.1 & 38.6 \\
RE \textsuperscript{3} \cite{hu2019retrieve} & 50.2 & 41.9 \\
Multi-passage \cite{wang2019multi} & 60.9 & 53.0 \\
Graph-retriever \cite{asai2019learning} & 63.8 & 56.5 \\
\hline
SPARTA & \textbf{66.5} & \textbf{59.3} \\

\Xhline{3\arrayrulewidth}

\textbf{OpenNQ} & \multicolumn{2}{c}{EM} \\ 
Model & Dev & Test \\ \hline
BERT + BM25 \cite{lee2018ranking} & 24.8 & 26.5 \\
Hard EM \cite{min2019discrete} & 28.8 & 28.1\\
ORQA\cite{lee2019latent}  & 31.3 & 33.3  \\
Graph-retriever \cite{asai2019learning} & 31.7 & 32.6 \\
\hline
SPARTA & \textbf{36.8} & \textbf{37.5} \\
\hline

\end{tabular}
\caption{Results on English Open SQuAD and NQ}
\label{tbl:openqa-en}
\end{table}

Table \ref{tbl:openqa-en} and \ref{tbl:openqa-zh} shows the SPARTA performance in OpenQA settings, tested in both English and Chinese datasets. Experimental results show that SPARTA retriever outperforms all existing models and obtains new state-of-the-art results on all four datasets. For OpenSQuAD and OpenNQ, SPARTA outperforms the previous best system~\cite{asai2019learning} by 2.7 absolute F1 points and 5.1 absolute EM points respectively. For OpenCMRC and OpenDRCD, SPARTA achieves a 15.3 and 6.7 absolute F1 points improvement over the previous best system~\cite{xie2020distant}.


Notably, the previous best system on OpenSQuAD and OpenNQ depends on sophisticated graph reasoning~\cite{asai2019learning}, whereas the proposed SPARTA system only uses single-hop ranker and require much less computation power. This suggests that for tasks that requires only single-hop reasoning, there is still big improvement room for better ranker-reader QA systems.

\begin{table}[]
\centering
\begin{tabular}{p{0.32\textwidth}p{0.05\textwidth}p{0.05\textwidth}} 
\hline
\textbf{OpenCMRC} &  &  \\ 
Model & F1 & EM \\ \hline
BERTserini\cite{xie2020distant} & 60.9 & 44.5 \\
BERTserini+DS \cite{xie2020distant} & 64.6 & 48.6 \\

\hline
SPARTA & \textbf{79.9} & \textbf{62.9} \\

\Xhline{3\arrayrulewidth}

\textbf{OpenDRCD} &  &  \\ 
Model & F1 & EM \\ \hline
BERTserini \cite{xie2020distant} & 65.0 & 50.7 \\
BERTserini+DS \cite{xie2020distant} & 67.7 & 55.4 \\
\hline
SPARTA & \textbf{74.6} & \textbf{63.1} \\
\hline

\end{tabular}
\caption{Results on Chinese Open CMRC and DRCD}
\label{tbl:openqa-zh}
\end{table}


%% file: reqa_exp.tex
We also consider Retrieval QA (ReQA), a sentence-level question answering task~\cite{ahmad2019reqa}. The candidate answer set contains every possible sentence from a text corpus and the system is expected to return a ranking of sentences given a query. The original ReQA only contains SQuAD and NQ. In this study, we extend ReQA to 11 different domains adapted from~\cite{fisch2019mrqa} to evaluate both \textit{in-domain performance} and \textit{out-of-domain generalization}. The details of the 11 ReQA domains are in Table~\ref{tbl:reqa_data} and Appendix.

The in-domain scenarios look at domains that have enough training data (see Table~\ref{tbl:reqa_data}). The models are trained on the training data and the evaluation is done on the test data. On the other hand, the out-of-domain scenarios evaluate systems' performance on test data from domains not included in the training, making it a zero-shot learning problem. There are two out-of-domain settings: (1) training data only contain SQuAD (2) training data contain only SQuAD and NQ. Evaluation is carried on all the domains to test systems' ability to generalize to unseen data distribution.

\begin{table}[ht]
\small
\begin{tabular}{p{0.28\textwidth}p{0.16\textwidth}} \hline
\textbf{Domain}                        & \textbf{Data Source}  \\ \hline
Has training data & \\ \hline
SQuAD~\cite{rajpurkar2016squad}          & Wikipedia                    \\
News~\cite{trischler2016newsqa}          & News                       \\
Trivia~\cite{joshi2017triviaqa}          & Web                        \\
NQ~\cite{kwiatkowski2019natural}         & Google Search                                           \\
Hotpot~\cite{yang2018hotpotqa}           & Wikipedia                       \\ \hline
Has no training data                    &                              \\ \hline
BioASQ~\cite{tsatsaronis2015overview}    & PubMed Documents                  \\
DROP~\cite{dua2019drop}                  & Wikipedia                       \\
DuoRC~\cite{saha2018duorc}               & Wikipedia+IMDB                       \\
RACE~\cite{lai2017race}                  & English Exam                       \\
RE~\cite{levy2017zero}                   & Wikipedia                       \\
Textbook~\cite{kembhavi2017you}          & K12 Textbook                  \\ \hline
\end{tabular}
\caption{11 corpora included in MultiReQA and their document sources. The top 5 domains contain training data and the bottom 6 domains only have test sets.}
\label{tbl:reqa_data}
\end{table}

For evaluation metrics, we use \textit{Mean Reciprocal Rank} (MRR) as the criteria. The competing baselines include:

\textbf{BM25}: a strong classic IR baseline that is difficult to beat~\cite{robertson2009probabilistic}.

\textbf{USE-QA}\footnote{\url{https://tfhub.dev/google/universal-sentence-encoder-multilingual-qa}}: universal sentence encoder trained for QA task by Google~\cite{yang2019multilingual}. USE-QA uses the dual-encoder architecture and it is trained on more than 900 million mined question-answer pairs with 16 different languages. 

\textbf{Poly-Encoder (Poly-Enc):} Poly Encoders improves the expressiveness of dual-encoders with two-level interaction~\cite{humeau2019poly}. We adapted the original dialog model for QA retrieval: two \textit{bert-base-uncased} models are used as the question and answer encoders. The answer encoder has 4 vector outputs.

%% file: reqa_results.tex
\subsection{In-domain Performance}
\begin{table}[]
\centering
\begin{tabular}{p{0.07\textwidth}|p{0.07\textwidth}p{0.07\textwidth}p{0.07\textwidth}p{0.07\textwidth}} \hline
\textbf{Data} & \textbf{BM25} & \textbf{USE-QA} & \textbf{Poly Enc} & \textbf{SPARTA (ours)}\\ \hline
SQuAD  & 58.0 & 62.5   & 64.6     & \textbf{78.5}   \\
News   & 19.4 & 26.2 & 28.3     & \textbf{46.6}  \\
Trivia & 29.0   & 41.2   & 39.5     & \textbf{55.5}   \\
NQ     & 19.7 & 58.2   & 69.9     & \textbf{77.1}   \\
HotPot & 23.9 & 25.5   & 51.8     & \textbf{63.8}  \\ \hline
Avg    & 30.0 &  42.7  &  50.8  & \textbf{64.3} \\ \hline
\end{tabular}
\caption{MRR comparison for the in-domain settings. The proposed SPARTA consistently outperform all the baseline models with large margin. BM25 and USE-QA are unsupervised and pre-trained respectively.}
\label{tbl:in-domain}
\end{table}

\begin{table*}[!ht]
\centering
\small
\begin{tabular}[width=16cm]{l|lllll|llllll|l} \hline
 \textbf{Model}   & \textbf{SQuAD} & \textbf{News}  & \textbf{Trivia} & \textbf{NQ}    & \textbf{HotPot} & \textbf{Bio}   & \textbf{DROP}  & \textbf{DuoRC} & \textbf{RACE}  & \textbf{RE}    & \textbf{Text}  & \textbf{Avg}   \\ \hline
 & \multicolumn{12}{c}{Unsupervised or pretrained}                                              \\ \hline
BM25      & 58.0 & 19.4 & 29.0  & 19.7  & 23.9  & 8.9 & 32.6 & 20.1 & 14.8 & 87.4 & 21.6 & 30.5 \\
USE-QA    & 62.5 & 26.2 & 41.2  & 58.2 & 25.5  & 7.7 & 31.9 & 20.8 & 25.6 & 84.8 & 26.4 & 37.4 \\ \hline
          & \multicolumn{12}{c}{Trained on SQuAD}                                              \\ \hline
PolyEnc & 64.6* & 22.2 & 35.9  & 57.6 & 26.5  & 9.1 & 32.6 & 25.4 & 24.7 & 88.3 & 26.0 & 37.5  \\
SPARTA    & 78.5* & \textbf{41.2} & 45.8  & 62.0 & \textbf{47.7}  & 14.5 & 37.2 & 35.9 & 29.7 & 96.0& 28.7 & 47.0 \\\hline
          & \multicolumn{12}{c}{Trained on SQuAD + NQ}                                              \\ \hline
PolyEnc & 63.9* & 19.8 & 36.9  & 69.7* & 29.6  & 8.8 & 30.7 & 19.6 & 25.2 & 72.8 & 24.6 & 36.5 \\
SPARTA    &\textbf{79.0}*  & 40.3  & \textbf{47.6} & \textbf{75.8}*  & 47.5 & \textbf{15.0} &\textbf{ 37.9} & \textbf{36.3} & \textbf{30.0} & \textbf{97.0} & \textbf{29.3} & \textbf{48.7} \\\hline
\end{tabular}
\caption{MRR comparison in the out-of-domain settings. The proposed SPARTA is able to achieve the best performance across all tasks, the only learning-based method that is able to consistently outperform BM25 with larger margin in new domains. Results with * are in-domain performance.}
\label{tbl:out-domain}
\end{table*}

Table~\ref{tbl:in-domain} shows the MRR results on the five datasets with in-domain training. SPARTA can achieve the best performance across all domains with a large main. In terms of average MRR across the five domains, SPARTA is 114.3\% better than BM25, 50.6\% better than USE-QA and 26.5\% better than Poly-Encoders. 

Two additional insights can be drawn from the results. First, BM-25 is a strong baseline and does not require training. It performs particularly well in domains that have a high-rate of word-overlapping between the answer and the questions. For example, SQuAD's questions are generated by crowd workers who look at the ground truth answer, while question data from NQ/News are generated by question makers who do not see the correct answer. BM25 works particularly well in SQuAD while performing the poorest in other datasets. Similar observations are also found in prior research~\cite{ahmad2019reqa}.

Second, the results in Table~\ref{tbl:in-domain} confirms our hypothesis on the importance of rich interaction between the answer and the questions. Both USE-QA and Poly Encoder use powerful transformers to encode the whole question and model word-order information in the queries. However, their performance is bounded by the simple dot-product interaction between the query and the answer. On the other hand, despite the fact that SPARTA does not model word-order information in the query, it is able to achieve a big performance gain compared to the baselines, confirming the effectiveness of the proposed token-level interaction method in Eq.~\ref{eq:term_match}.

\subsection{Out-of-domain Generalization}
Table~\ref{tbl:out-domain} summarized the results for out-of-domain performance comparison. SPARTA trained only on SQuAD outperforms the baselines, achieving 54.1\% gain compared to BM25, 26.7\% gain compared to USE-QA and 25.3\% gain compared to Poly-Encoders in terms of average MRR across 11 different datasets. When SPARTA is trained on SQuAD+NQ, an additional 1.7 MRR improvement is gained compared to SPARTA-SQuAD. 



We can observe that Poly-Encoder is able to achieve similar in-domain performance for the domains that are included in the training. However, its performance decreases significantly in new domains, a 25.0\% drop compared to its full performance for Poly-Encoder that is trained on SQuAD and 29.2\% drop when it's trained on SQuAD+NQ. 

Meanwhile, SPARTA generalizes its knowledge from the training data much better to new domains. When trained on SQuAD, its performance on News, Trivia, NQ, and HotPot is only 19.2\% lower than the full performance and 18.3\% drop when it's trained on SQuAD+NQ. Also, we note that SPARTA's zero-shot performance on News (MRR=41.2) and Trivia (MRR=45.8) is even better than the full performance of Poly-Encoder (News MRR=28.3 and Trivia MRR=39.5).






%% file: analysis.tex
\subsection{Interpreting Sparse Representations}
One common limitation of deep neural network models is poor interpretability.  Take dense distributed vector representation for example, one cannot directly make sense of each dimension and has to use dimension reduction and visualization methods, e.g. TSNE~\cite{maaten2008visualizing}. On the contrary, the resulting SPARTA index is straightforward to interpret due to its sparse nature. Specifically, we can understand a SPARTA vector by reading the top K words with non-zero $f(t, (a, c))$, since these terms have the greatest impact to the final ranking score. 

\begin{table*}[!ht]
    \small
    \centering 
    \begin{tabular}[width=16cm]{p{0.47\textwidth}|p{0.47\textwidth}} \hline
    \textbf{Answer $(a, c)$ }  &  \textbf{Top terms} \\ \hline
    \textbf{Google was founded in September 1998} by Larry Page and Sergey Brin while they were Ph.D. students at Stanford University in California. & google, when, founded, page, stanford, sergey, larry, founding, established, did, 1998, was, year, formed ... \\ \hline
    Yellowstone National Park is an American national park located in the western United States, \textbf{with parts in Wyoming, Montana and Idaho.} & montana,  yellowstone, wyoming, idaho, park, where, national, western, american, us, utah ... \\ \hline 
    \textbf{William Henry Gates is an American business magnate, software developer, investor, and philanthropist.} He is best known as the co-founder of Microsoft Corporation. &  who, gates, investors, magnate, developer, microsoft, philanthropist, benefactor, investors, ...  \\ \hline
    \textbf{Question answering (QA)} is a computer science discipline within the fields of information retrieval and natural language processing (NLP). &  answering, question, q, computer, information,, retrieval,language, natural, human, nl, science, ...  \\ \hline
    \end{tabular}
    \caption{Top-k terms predicted by SPARTA. The text in bold is the answer sentence and the text surrounded it is encoded as its context. Each answer sentence has around 1600 terms with non-zero scores.}
    \label{tbl:example}
\end{table*}

Table~\ref{tbl:example} shows some example outputs. It is not hard to note that the generated terms for each answer sentence is highly relevant to both $a$ and $c$, and contains not keywords that appeared in the answer, but also include terms that are potentially in the query but never appear in the answer itself. Two experts manually inspect the outputs for 500 $(a, c)$ data points from Wikipedia, and we summarize the following four major categories of terms that are predicted by SPARTA.  

\textbf{Conversational search understanding}: the third row is an example. ``Who'' appears to the top term, showing it learns Bill Gates is a person so that it's likely to match with ``Who'' questions. 

\textbf{Keyword identification}: terms such as ``gates, google, magnate, yellowstone'' have high scores in the generated vector, showing that SPARTA learns which words are important in the answer.


\textbf{Synonyms and Common Sense}: ``benefactor, investors'' are examples of synonyms. Also even though ``Utah'' does not appear in the answer, it is predicted as an important term, showing that SPARTA leverages the world-knowledge from a pretrained language model and knows Yellowstone is related to Utah.


\subsection{Sparsity vs. Performance}
\begin{table}[]
\centering
\begin{tabular}{l|ll|ll} \hline
\textbf{Top-K} & \multicolumn{2}{c|}{\textbf{SQuAD}} & \multicolumn{2}{c}{\textbf{NQ}}     \\ 
                       & \textbf{MRR}       & \textbf{R@1}       & \textbf{MRR}   &  \textbf{R@1}  \\ \hline

50    & 69.5 & 61.3 & 63.2 & 52.5 \\
100   & 72.3 & 64.4 & 65.6 & 55.7 \\
500   & 76.9 & 69.4 & 74.4 & 64.3 \\
1000  & 78.2 & 70.8 & 75.5 & 65.6 \\
1500  & 78.6 & 71.2 & 75.7 &  65.7 \\
2000  & 78.9 & 71.4 & 75.9 & 66.0 \\
Full  & 79.0 & 71.6 & 75.8 & 66.0 \\ \hline
\end{tabular}
\caption{Performance on ReQA task with varying sparsity. SPARTA outperforms all baselines with top-50 terms on SQuAD, and with top-500 terms on NQ.}
\label{tbl:sparsity}
\end{table}

Sparsity not only provides interpretability, but also offers flexibility to balance the trade-off of memory footprint vs. performance. When there are memory constraints on the vector size, the SPARTA vector can be easily reduced by only keeping the top-K important terms. Table~\ref{tbl:sparsity} shows performance on SQuAD and NQ with varying K. The resulting sparse vector representation is very robust to smaller K. When only keeping the top 50 terms in each answer vector, SPARTA achieves 69.5 MRR, a better score than all baselines with only $1.6\%$ memory footprint compared to Poly-Encoders (768 x 4 dimension). NQ dataset is more challenging and requires more terms. SPARTA achieves a close to the best performance with top-500 terms.

%% file: emnlp2020.bbl
\begin{thebibliography}{55}
\expandafter\ifx\csname natexlab\endcsname\relax\def\natexlab#1{#1}\fi

\bibitem[{Ahmad et~al.(2019)Ahmad, Constant, Yang, and Cer}]{ahmad2019reqa}
Amin Ahmad, Noah Constant, Yinfei Yang, and Daniel Cer. 2019.
\newblock Reqa: An evaluation for end-to-end answer retrieval models.
\newblock \emph{arXiv preprint arXiv:1907.04780}.

\bibitem[{Asai et~al.(2019)Asai, Hashimoto, Hajishirzi, Socher, and
  Xiong}]{asai2019learning}
Akari Asai, Kazuma Hashimoto, Hannaneh Hajishirzi, Richard Socher, and Caiming
  Xiong. 2019.
\newblock Learning to retrieve reasoning paths over wikipedia graph for
  question answering.
\newblock \emph{arXiv preprint arXiv:1911.10470}.

\bibitem[{Berant et~al.(2013)Berant, Chou, Frostig, and
  Liang}]{berant2013semantic}
Jonathan Berant, Andrew Chou, Roy Frostig, and Percy Liang. 2013.
\newblock Semantic parsing on freebase from question-answer pairs.
\newblock In \emph{Proceedings of the 2013 conference on empirical methods in
  natural language processing}, pages 1533--1544.

\bibitem[{Berant and Liang(2014)}]{berant2014semantic}
Jonathan Berant and Percy Liang. 2014.
\newblock Semantic parsing via paraphrasing.
\newblock In \emph{Proceedings of the 52nd Annual Meeting of the Association
  for Computational Linguistics (Volume 1: Long Papers)}, pages 1415--1425.

\bibitem[{Bollacker et~al.(2008)Bollacker, Evans, Paritosh, Sturge, and
  Taylor}]{bollacker2008freebase}
Kurt Bollacker, Colin Evans, Praveen Paritosh, Tim Sturge, and Jamie Taylor.
  2008.
\newblock Freebase: a collaboratively created graph database for structuring
  human knowledge.
\newblock In \emph{Proceedings of the 2008 ACM SIGMOD international conference
  on Management of data}, pages 1247--1250.

\bibitem[{Chang et~al.(2020)Chang, Yu, Chang, Yang, and Kumar}]{chang2020pre}
Wei-Cheng Chang, Felix~X Yu, Yin-Wen Chang, Yiming Yang, and Sanjiv Kumar.
  2020.
\newblock Pre-training tasks for embedding-based large-scale retrieval.
\newblock \emph{arXiv preprint arXiv:2002.03932}.

\bibitem[{Chen et~al.(2017)Chen, Fisch, Weston, and Bordes}]{chen2017reading}
Danqi Chen, Adam Fisch, Jason Weston, and Antoine Bordes. 2017.
\newblock Reading {Wikipedia} to answer open-domain questions.
\newblock In \emph{Association for Computational Linguistics (ACL)}.

\bibitem[{Chidambaram et~al.(2018)Chidambaram, Yang, Cer, Yuan, Sung, Strope,
  and Kurzweil}]{chidambaram2018learning}
Muthuraman Chidambaram, Yinfei Yang, Daniel Cer, Steve Yuan, Yun-Hsuan Sung,
  Brian Strope, and Ray Kurzweil. 2018.
\newblock Learning cross-lingual sentence representations via a multi-task
  dual-encoder model.
\newblock \emph{arXiv preprint arXiv:1810.12836}.

\bibitem[{Clark and Gardner(2017)}]{clark2017simple}
Christopher Clark and Matt Gardner. 2017.
\newblock Simple and effective multi-paragraph reading comprehension.
\newblock \emph{arXiv preprint arXiv:1710.10723}.

\bibitem[{Cui et~al.(2020)Cui, Che, Liu, Qin, Wang, and
  Hu}]{cui-etal-2020-revisiting}
Yiming Cui, Wanxiang Che, Ting Liu, Bing Qin, Shijin Wang, and Guoping Hu.
  2020.
\newblock Revisiting pre-trained models for chinese natural language
  processing.
\newblock In \emph{Findings of EMNLP}. Association for Computational
  Linguistics.

\bibitem[{Cui et~al.(2018)Cui, Liu, Che, Xiao, Chen, Ma, Wang, and
  Hu}]{cui2018span}
Yiming Cui, Ting Liu, Wanxiang Che, Li~Xiao, Zhipeng Chen, Wentao Ma, Shijin
  Wang, and Guoping Hu. 2018.
\newblock A span-extraction dataset for chinese machine reading comprehension.
\newblock \emph{arXiv preprint arXiv:1810.07366}.

\bibitem[{Devlin et~al.(2018)Devlin, Chang, Lee, and
  Toutanova}]{devlin2018bert}
Jacob Devlin, Ming-Wei Chang, Kenton Lee, and Kristina Toutanova. 2018.
\newblock Bert: Pre-training of deep bidirectional transformers for language
  understanding.
\newblock \emph{arXiv preprint arXiv:1810.04805}.

\bibitem[{Dhingra et~al.(2020)Dhingra, Zaheer, Balachandran, Neubig,
  Salakhutdinov, and Cohen}]{dhingra2020differentiable}
Bhuwan Dhingra, Manzil Zaheer, Vidhisha Balachandran, Graham Neubig, Ruslan
  Salakhutdinov, and William~W Cohen. 2020.
\newblock Differentiable reasoning over a virtual knowledge base.
\newblock \emph{arXiv preprint arXiv:2002.10640}.

\bibitem[{Dua et~al.(2019)Dua, Wang, Dasigi, Stanovsky, Singh, and
  Gardner}]{dua2019drop}
Dheeru Dua, Yizhong Wang, Pradeep Dasigi, Gabriel Stanovsky, Sameer Singh, and
  Matt Gardner. 2019.
\newblock Drop: A reading comprehension benchmark requiring discrete reasoning
  over paragraphs.
\newblock \emph{arXiv preprint arXiv:1903.00161}.

\bibitem[{Fader et~al.(2014)Fader, Zettlemoyer, and Etzioni}]{fader2014open}
Anthony Fader, Luke Zettlemoyer, and Oren Etzioni. 2014.
\newblock Open question answering over curated and extracted knowledge bases.
\newblock In \emph{Proceedings of the 20th ACM SIGKDD international conference
  on Knowledge discovery and data mining}, pages 1156--1165.

\bibitem[{Ferrucci et~al.(2010)Ferrucci, Brown, Chu-Carroll, Fan, Gondek,
  Kalyanpur, Lally, Murdock, Nyberg, Prager et~al.}]{ferrucci2010building}
David Ferrucci, Eric Brown, Jennifer Chu-Carroll, James Fan, David Gondek,
  Aditya~A Kalyanpur, Adam Lally, J~William Murdock, Eric Nyberg, John Prager,
  et~al. 2010.
\newblock Building watson: An overview of the deepqa project.
\newblock \emph{AI magazine}, 31(3):59--79.

\bibitem[{Fisch et~al.(2019)Fisch, Talmor, Jia, Seo, Choi, and
  Chen}]{fisch2019mrqa}
Adam Fisch, Alon Talmor, Robin Jia, Minjoon Seo, Eunsol Choi, and Danqi Chen.
  2019.
\newblock Mrqa 2019 shared task: Evaluating generalization in reading
  comprehension.
\newblock \emph{arXiv preprint arXiv:1910.09753}.

\bibitem[{Guo et~al.(2016)Guo, Fan, Ai, and Croft}]{guo2016deep}
Jiafeng Guo, Yixing Fan, Qingyao Ai, and W~Bruce Croft. 2016.
\newblock A deep relevance matching model for ad-hoc retrieval.
\newblock In \emph{Proceedings of the 25th ACM International on Conference on
  Information and Knowledge Management}, pages 55--64.

\bibitem[{Henderson et~al.(2019)Henderson, Casanueva, Mrk{\v{s}}i{\'c}, Su,
  Vuli{\'c} et~al.}]{henderson2019convert}
Matthew Henderson, I{\~n}igo Casanueva, Nikola Mrk{\v{s}}i{\'c}, Pei-Hao Su,
  Ivan Vuli{\'c}, et~al. 2019.
\newblock Convert: Efficient and accurate conversational representations from
  transformers.
\newblock \emph{arXiv preprint arXiv:1911.03688}.

\bibitem[{Hu et~al.(2019)Hu, Peng, Huang, and Li}]{hu2019retrieve}
Minghao Hu, Yuxing Peng, Zhen Huang, and Dongsheng Li. 2019.
\newblock Retrieve, read, rerank: Towards end-to-end multi-document reading
  comprehension.
\newblock \emph{arXiv preprint arXiv:1906.04618}.

\bibitem[{Humeau et~al.(2019)Humeau, Shuster, Lachaux, and
  Weston}]{humeau2019poly}
Samuel Humeau, Kurt Shuster, Marie-Anne Lachaux, and Jason Weston. 2019.
\newblock Poly-encoders: Transformer architectures and pre-training strategies
  for fast and accurate multi-sentence scoring.
\newblock \emph{CoRR abs/1905.01969. External Links: Link Cited by}, 2:2--2.

\bibitem[{Joshi et~al.(2020)Joshi, Chen, Liu, Weld, Zettlemoyer, and
  Levy}]{joshi2020spanbert}
Mandar Joshi, Danqi Chen, Yinhan Liu, Daniel~S Weld, Luke Zettlemoyer, and Omer
  Levy. 2020.
\newblock Spanbert: Improving pre-training by representing and predicting
  spans.
\newblock \emph{Transactions of the Association for Computational Linguistics},
  8:64--77.

\bibitem[{Joshi et~al.(2017)Joshi, Choi, Weld, and
  Zettlemoyer}]{joshi2017triviaqa}
Mandar Joshi, Eunsol Choi, Daniel~S Weld, and Luke Zettlemoyer. 2017.
\newblock Triviaqa: A large scale distantly supervised challenge dataset for
  reading comprehension.
\newblock \emph{arXiv preprint arXiv:1705.03551}.

\bibitem[{Karpukhin et~al.(2020)Karpukhin, O{\u{g}}uz, Min, Wu, Edunov, Chen,
  and Yih}]{karpukhin2020dense}
Vladimir Karpukhin, Barlas O{\u{g}}uz, Sewon Min, Ledell Wu, Sergey Edunov,
  Danqi Chen, and Wen-tau Yih. 2020.
\newblock Dense passage retrieval for open-domain question answering.
\newblock \emph{arXiv preprint arXiv:2004.04906}.

\bibitem[{Kembhavi et~al.(2017)Kembhavi, Seo, Schwenk, Choi, Farhadi, and
  Hajishirzi}]{kembhavi2017you}
Aniruddha Kembhavi, Minjoon Seo, Dustin Schwenk, Jonghyun Choi, Ali Farhadi,
  and Hannaneh Hajishirzi. 2017.
\newblock Are you smarter than a sixth grader? textbook question answering for
  multimodal machine comprehension.
\newblock In \emph{Proceedings of the IEEE Conference on Computer Vision and
  Pattern Recognition}, pages 4999--5007.

\bibitem[{Kingma and Ba(2014)}]{kingma2014adam}
Diederik~P Kingma and Jimmy Ba. 2014.
\newblock Adam: A method for stochastic optimization.
\newblock \emph{arXiv preprint arXiv:1412.6980}.

\bibitem[{Kwiatkowski et~al.(2019)Kwiatkowski, Palomaki, Redfield, Collins,
  Parikh, Alberti, Epstein, Polosukhin, Devlin, Lee
  et~al.}]{kwiatkowski2019natural}
Tom Kwiatkowski, Jennimaria Palomaki, Olivia Redfield, Michael Collins, Ankur
  Parikh, Chris Alberti, Danielle Epstein, Illia Polosukhin, Jacob Devlin,
  Kenton Lee, et~al. 2019.
\newblock Natural questions: a benchmark for question answering research.
\newblock \emph{Transactions of the Association for Computational Linguistics},
  7:453--466.

\bibitem[{Lai et~al.(2017)Lai, Xie, Liu, Yang, and Hovy}]{lai2017race}
Guokun Lai, Qizhe Xie, Hanxiao Liu, Yiming Yang, and Eduard Hovy. 2017.
\newblock Race: Large-scale reading comprehension dataset from examinations.
\newblock \emph{arXiv preprint arXiv:1704.04683}.

\bibitem[{Lee et~al.(2020)Lee, Seo, Hajishirzi, and
  Kang}]{Lee2020ContextualizedSR}
Jinhyuk Lee, Minjoon Seo, Hannaneh Hajishirzi, and Jaewoo Kang. 2020.
\newblock Contextualized sparse representations for real-time open-domain
  question answering.
\newblock \emph{arXiv: Computation and Language}.

\bibitem[{Lee et~al.(2018)Lee, Yun, Kim, Ko, and Kang}]{lee2018ranking}
Jinhyuk Lee, Seongjun Yun, Hyunjae Kim, Miyoung Ko, and Jaewoo Kang. 2018.
\newblock Ranking paragraphs for improving answer recall in open-domain
  question answering.
\newblock \emph{arXiv preprint arXiv:1810.00494}.

\bibitem[{Lee et~al.(2019)Lee, Chang, and Toutanova}]{lee2019latent}
Kenton Lee, Ming-Wei Chang, and Kristina Toutanova. 2019.
\newblock Latent retrieval for weakly supervised open domain question
  answering.
\newblock \emph{arXiv preprint arXiv:1906.00300}.

\bibitem[{Levy et~al.(2017)Levy, Seo, Choi, and Zettlemoyer}]{levy2017zero}
Omer Levy, Minjoon Seo, Eunsol Choi, and Luke Zettlemoyer. 2017.
\newblock Zero-shot relation extraction via reading comprehension.
\newblock \emph{arXiv preprint arXiv:1706.04115}.

\bibitem[{Lewis and Fan(2018)}]{lewis2018generative}
Mike Lewis and Angela Fan. 2018.
\newblock Generative question answering: Learning to answer the whole question.

\bibitem[{Maaten and Hinton(2008)}]{maaten2008visualizing}
Laurens van~der Maaten and Geoffrey Hinton. 2008.
\newblock Visualizing data using t-sne.
\newblock \emph{Journal of machine learning research}, 9(Nov):2579--2605.

\bibitem[{Manning et~al.(2008)Manning, Raghavan, and
  Sch{\"u}tze}]{manning2008introduction}
Christopher~D Manning, Prabhakar Raghavan, and Hinrich Sch{\"u}tze. 2008.
\newblock \emph{Introduction to information retrieval}.
\newblock Cambridge university press.

\bibitem[{McCandless et~al.(2010)McCandless, Hatcher, Gospodneti{\'c}, and
  Gospodneti{\'c}}]{mccandless2010lucene}
Michael McCandless, Erik Hatcher, Otis Gospodneti{\'c}, and O~Gospodneti{\'c}.
  2010.
\newblock \emph{Lucene in action}, volume~2.
\newblock Manning Greenwich.

\bibitem[{Min et~al.(2019)Min, Chen, Hajishirzi, and
  Zettlemoyer}]{min2019discrete}
Sewon Min, Danqi Chen, Hannaneh Hajishirzi, and Luke Zettlemoyer. 2019.
\newblock A discrete hard em approach for weakly supervised question answering.
\newblock \emph{arXiv preprint arXiv:1909.04849}.

\bibitem[{Min et~al.(2018)Min, Zhong, Socher, and Xiong}]{min2018efficient}
Sewon Min, Victor Zhong, Richard Socher, and Caiming Xiong. 2018.
\newblock Efficient and robust question answering from minimal context over
  documents.
\newblock \emph{arXiv preprint arXiv:1805.08092}.

\bibitem[{Nogueira et~al.(2019)Nogueira, Lin, and
  Epistemic}]{nogueira2019doc2query}
Rodrigo Nogueira, Jimmy Lin, and AI~Epistemic. 2019.
\newblock From doc2query to doctttttquery.

\bibitem[{Rajpurkar et~al.(2016)Rajpurkar, Zhang, Lopyrev, and
  Liang}]{rajpurkar2016squad}
Pranav Rajpurkar, Jian Zhang, Konstantin Lopyrev, and Percy Liang. 2016.
\newblock Squad: 100,000+ questions for machine comprehension of text.
\newblock \emph{arXiv preprint arXiv:1606.05250}.

\bibitem[{Robertson et~al.(2009)Robertson, Zaragoza
  et~al.}]{robertson2009probabilistic}
Stephen Robertson, Hugo Zaragoza, et~al. 2009.
\newblock The probabilistic relevance framework: Bm25 and beyond.
\newblock \emph{Foundations and Trends{\textregistered} in Information
  Retrieval}, 3(4):333--389.

\bibitem[{Saha et~al.(2018)Saha, Aralikatte, Khapra, and
  Sankaranarayanan}]{saha2018duorc}
Amrita Saha, Rahul Aralikatte, Mitesh~M Khapra, and Karthik Sankaranarayanan.
  2018.
\newblock Duorc: Towards complex language understanding with paraphrased
  reading comprehension.
\newblock \emph{arXiv preprint arXiv:1804.07927}.

\bibitem[{Seo et~al.(2016)Seo, Kembhavi, Farhadi, and
  Hajishirzi}]{seo2016bidirectional}
Minjoon Seo, Aniruddha Kembhavi, Ali Farhadi, and Hannaneh Hajishirzi. 2016.
\newblock Bidirectional attention flow for machine comprehension.
\newblock \emph{arXiv preprint arXiv:1611.01603}.

\bibitem[{Seo et~al.(2019)Seo, Lee, Kwiatkowski, Parikh, Farhadi, and
  Hajishirzi}]{seo2019real}
Minjoon Seo, Jinhyuk Lee, Tom Kwiatkowski, Ankur Parikh, Ali Farhadi, and
  Hannaneh Hajishirzi. 2019.
\newblock Real-time open-domain question answering with dense-sparse phrase
  index.
\newblock In \emph{Proceedings of the 57th Annual Meeting of the Association
  for Computational Linguistics}, pages 4430--4441.

\bibitem[{Shao et~al.(2018)Shao, Liu, Lai, Tseng, and Tsai}]{shao2018drcd}
Chih~Chieh Shao, Trois Liu, Yuting Lai, Yiying Tseng, and Sam Tsai. 2018.
\newblock Drcd: a chinese machine reading comprehension dataset.
\newblock \emph{arXiv preprint arXiv:1806.00920}.

\bibitem[{Shrivastava and Li(2014)}]{shrivastava2014asymmetric}
Anshumali Shrivastava and Ping Li. 2014.
\newblock Asymmetric lsh (alsh) for sublinear time maximum inner product search
  (mips).
\newblock In \emph{Advances in Neural Information Processing Systems}, pages
  2321--2329.

\bibitem[{Trischler et~al.(2016)Trischler, Wang, Yuan, Harris, Sordoni,
  Bachman, and Suleman}]{trischler2016newsqa}
Adam Trischler, Tong Wang, Xingdi~(Eric) Yuan, Justin~D. Harris, Alessandro
  Sordoni, Philip Bachman, and Kaheer Suleman. 2016.
\newblock \href
  {https://www.microsoft.com/en-us/research/publication/newsqa-machine-comprehension-dataset/}
  {Newsqa: A machine comprehension dataset}.

\bibitem[{Tsatsaronis et~al.(2015)Tsatsaronis, Balikas, Malakasiotis, Partalas,
  Zschunke, Alvers, Weissenborn, Krithara, Petridis, Polychronopoulos
  et~al.}]{tsatsaronis2015overview}
George Tsatsaronis, Georgios Balikas, Prodromos Malakasiotis, Ioannis Partalas,
  Matthias Zschunke, Michael~R Alvers, Dirk Weissenborn, Anastasia Krithara,
  Sergios Petridis, Dimitris Polychronopoulos, et~al. 2015.
\newblock An overview of the bioasq large-scale biomedical semantic indexing
  and question answering competition.
\newblock \emph{BMC bioinformatics}, 16(1):138.

\bibitem[{Wang et~al.(2018)Wang, Yu, Guo, Wang, Klinger, Zhang, Chang, Tesauro,
  Zhou, and Jiang}]{wang2018r}
Shuohang Wang, Mo~Yu, Xiaoxiao Guo, Zhiguo Wang, Tim Klinger, Wei Zhang, Shiyu
  Chang, Gerry Tesauro, Bowen Zhou, and Jing Jiang. 2018.
\newblock R 3: Reinforced ranker-reader for open-domain question answering.
\newblock In \emph{Thirty-Second AAAI Conference on Artificial Intelligence}.

\bibitem[{Wang et~al.(2019)Wang, Ng, Ma, Nallapati, and Xiang}]{wang2019multi}
Zhiguo Wang, Patrick Ng, Xiaofei Ma, Ramesh Nallapati, and Bing Xiang. 2019.
\newblock Multi-passage bert: A globally normalized bert model for open-domain
  question answering.
\newblock \emph{arXiv preprint arXiv:1908.08167}.

\bibitem[{Xie et~al.(2020)Xie, Yang, Tan, Xiong, Yuan, Huai, Li, and
  Lin}]{xie2020distant}
Yuqing Xie, Wei Yang, Luchen Tan, Kun Xiong, Nicholas~Jing Yuan, Baoxing Huai,
  Ming Li, and Jimmy Lin. 2020.
\newblock Distant supervision for multi-stage fine-tuning in retrieval-based
  question answering.
\newblock In \emph{Proceedings of The Web Conference 2020}, pages 2934--2940.

\bibitem[{Xiong et~al.(2017)Xiong, Dai, Callan, Liu, and Power}]{xiong2017end}
Chenyan Xiong, Zhuyun Dai, Jamie Callan, Zhiyuan Liu, and Russell Power. 2017.
\newblock End-to-end neural ad-hoc ranking with kernel pooling.
\newblock In \emph{Proceedings of the 40th International ACM SIGIR conference
  on research and development in information retrieval}, pages 55--64.

\bibitem[{Yang et~al.(2019{\natexlab{a}})Yang, Xie, Lin, Li, Tan, Xiong, Li,
  and Lin}]{yang2019end}
Wei Yang, Yuqing Xie, Aileen Lin, Xingyu Li, Luchen Tan, Kun Xiong, Ming Li,
  and Jimmy Lin. 2019{\natexlab{a}}.
\newblock End-to-end open-domain question answering with bertserini.
\newblock \emph{arXiv preprint arXiv:1902.01718}.

\bibitem[{Yang et~al.(2019{\natexlab{b}})Yang, Cer, Ahmad, Guo, Law, Constant,
  Abrego, Yuan, Tar, Sung et~al.}]{yang2019multilingual}
Yinfei Yang, Daniel Cer, Amin Ahmad, Mandy Guo, Jax Law, Noah Constant,
  Gustavo~Hernandez Abrego, Steve Yuan, Chris Tar, Yun-Hsuan Sung, et~al.
  2019{\natexlab{b}}.
\newblock Multilingual universal sentence encoder for semantic retrieval.
\newblock \emph{arXiv preprint arXiv:1907.04307}.

\bibitem[{Yang et~al.(2018)Yang, Qi, Zhang, Bengio, Cohen, Salakhutdinov, and
  Manning}]{yang2018hotpotqa}
Zhilin Yang, Peng Qi, Saizheng Zhang, Yoshua Bengio, William~W Cohen, Ruslan
  Salakhutdinov, and Christopher~D Manning. 2018.
\newblock Hotpotqa: A dataset for diverse, explainable multi-hop question
  answering.
\newblock \emph{arXiv preprint arXiv:1809.09600}.

\end{thebibliography}
